# Dynamic-Pix2Pix: Noise Injected cGAN for Modeling Input and Target Domain Joint Distributions with Limited Training Data

Mohammadreza Naderi[1], Nader Karimi[1], Ali Emami[1], Shahram Shirani[2], Shadrokh Samavi[1,2,3]

[1]Department of Electrical and Computer Engineering, Isfahan University of Technology, 84156-83111, Iran
[2]Department of Electrical and Computer Engineering, McMaster University, L8S 4L8, Canada
[3]Computer Science Department, Seattle University, 98122, USA

## Abstract

Learning to translate images from a source to a target domain with applications such as converting simple line drawing to oil painting has attracted significant attention. The quality of translated images is directly related to two crucial issues. First, the consistency of the output distribution with that of the target is essential. Second, the generated output should have a high correlation with the input. Conditional Generative Adversarial Networks, cGANs, are the most common models for translating images. The performance of a cGAN drops when we use a limited training dataset. In this work, we increase the Pix2Pix (a form of cGAN) target distribution modeling ability with the help of dynamic neural network theory. Our model has two learning cycles. The model learns the correlation between input and ground truth in the first cycle. Then, the model's architecture is refined in the second cycle to learn the target distribution from noise input. These processes are executed in each iteration of the training procedure. Helping the cGAN learn the target distribution from noise input results in a better model generalization during the test time and allows the model to fit almost perfectly to the target domain distribution. As a result, our model surpasses the Pix2Pix model in segmenting HC18 and Montgomery's chest x-ray images. Both qualitative and Dice scores show the superiority of our model. Although our proposed method does not use thousand of additional data for pretraining, it produces comparable results for the in and out-domain generalization compared to the state-of-the-art methods.

## 1. Introduction

Generative adversarial networks (GANs) can model the distribution of images when a large dataset from that distribution is available. However, as explained in [1], the simple GAN architectures need at least 100k target domain images to learn that target domain distribution. Conditional generative adversarial networks (cGANs) attempted to expand the GAN's idea to the image translation problems such as image segmentation, image inpainting, and image composition. However, acquiring an extensive dataset could be impossible in some cases, especially in medical fields, where it could be time-consuming and costly. Therefore, cGANs learn the joint distribution of input and target images instead of learning the target domain distribution.

Noise is omitted from the cGANs architectures because they do not have enough images to learn a complete interpretation of the target domain. Thus, insufficient data pushes cGANs to fit generated images to the joint target and input domain distribution. We want cGAN to learn to generate images that precisely fit the target domain distribution. The lack of concentration on the target domain could cause the cGAN to violate the target domain's key factors. We can see such outcomes during testing when the input data differs from the training distribution. For instance, in the Pix2Pix [2] architecture, there are two losses: pixel-wise loss and adversarial loss. When we have a small dataset, the pixel-wise loss has a much higher impact on the initial learning steps to help the generator to converge faster and make more sensible outputs. After the initial learning steps, the two losses struggle to produce results that fit the joint target and input distribution while satisfying the pixel-wise loss. This prevents the generator from generating a large number of possible fake images. Thus, the discriminator fails to learn a complete target distribution. In [3], the authors showed that learning an entire joint input and target distribution in the presence of thousands of images improves image translation ability and generalizability. However, they do not consider the cases of limited training data. We want to see if changing the Pix2Pix architecture could help the network thoroughly learn the target domain distribution using limited training data. Furthermore, we want to determine what elements in the GAN's architecture prevent the network from learning the target distribution from limited data.

### 1.1 Motivation

U-Net-like architectures, such as [4-6], learn to construct the output based on the extracted input features. Conditional GAN utilizes the same architecture as U-Net-like architectures for the generator. It also benefits from a discriminator network that helps the generator better fit the outputs to the target distribution. Thus, unlike U-Net-like architectures, cGANs generalize better on unseen data during the test time. However, there is still a downfall in the cGANs architecture. The discriminator cannot oblige the generator to construct an output that fits the target domain for unseen input images.

Since the input images passed to the network are limited and do not represent the entire space of the input domain, the generator does not construct most of the possible output images. Thus, the backpropagated gradients are not comprehensive enough when the generated images are fed as fake cases to the discriminator. Hence, the generator does not thoroughly learn the target distribution. Therefore, the generated output does not fit well with the target distribution for all seen and unseen images. Noise could represent all possible inputs that we can feed to the network. However, we cannot feed noise directly to cGAN's generator architecture as an input to generate all possible outputs. We will discuss the reasons in detail below.

For the same reason, we do not use U-Net-like architectures in GANs [7-11], in which we want to learn target distribution from noise. Instead, noise is fed to the generator part of a GAN to play the role of a low-dimensional code vector which the high-dimensional target domain must map to it. However, the noise vector usually has less than 1024 elements [7-11], giving the generator little information. Hence, during the training phase, the network realizes that it cannot extract enough information from the noise vector to make a high correlation between target images and the noise input. Since the generator can not extract meaningful information from the noise, it backpropagates most of the target domain features and treats the noise as a code. Every time the noise changes, the network realizes what output must be generated. As a result, the image space is mapped to the noise space by the generator.

In the upcoming scenarios, we consider a U-Net-like architecture as a generator for a simple GAN (instead of the typical decoder-like architecture) to see the effect of the noise on learning the desired distribution. In each case, we train the model to learn to segment the Montgomery chest x-ray dataset. We explain the architectural changes. Test results of each architecture are also shown to demonstrate the ability to model the distribution of labels. We can consider the following scenarios:

**a) Feeding noise to the generator:**
If we feed a noise-like image to the U-Net-like architectures, the noise will be of a high dimension. The network treats it as a normal image and tries extracting features from that noise, as shown in Figure 1(a). The generator attempts to relate the target images to the noise images; i.e., it attempts to segment the desired part from the noise image and does not treat it as a code to map the target distribution to it. We can explain the reason with the upcoming example: For a U-Net architecture [4] that takes a 256×256 image as input, feeding a noise implies that the noise has $256^2$ independent elements. Since the noise carries too much information to the U-Net, the architecture attempts to process it to construct output based on them. This will lead the network to learn a high correlation between noise and target images. In simple words, it attempts to construct the outputs based on noise features (like regular tasks where U-Net segments an image), and it fails to learn the target distribution as we want.

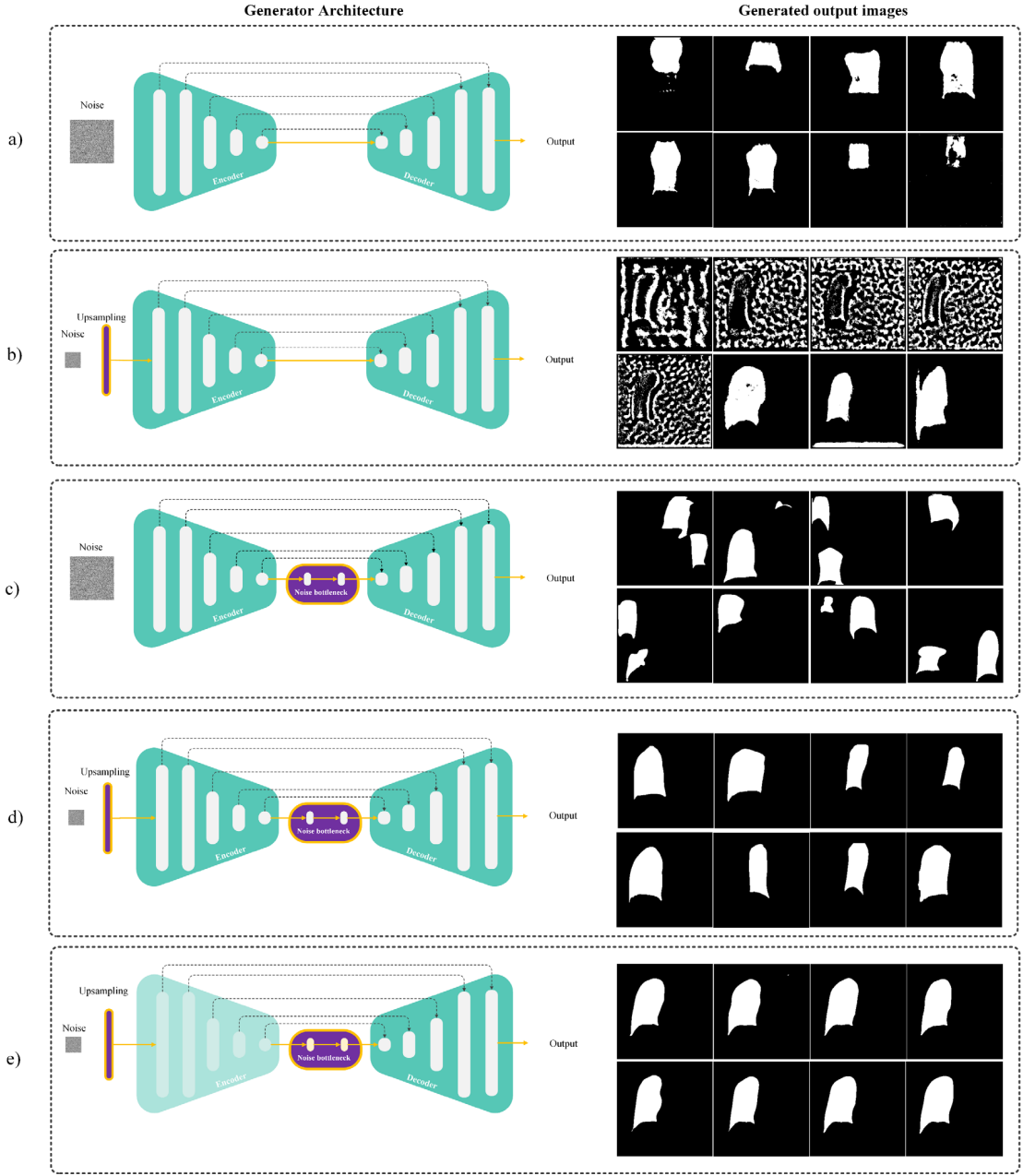

Figure 1: The behavior of the Pix2Pix to different noise injection scenarios to learn Montgomery ground truths distribution. a) Using noise with 256×256 size as input. b) Using low dimensional noise (4×4), upsample it to 256×256 and feed it to the generator as input. c) Using high-dimensional noise (256×256) as input and using a narrower bottleneck to the generator. d) Using a low-dimensional (4×4) and narrower bottleneck simultaneously. e) Freeze the encoder part of the generator and use the (d) setup. (Only the generator architecture is changed in each step. Thus, the discriminator architecture presents in Figure 2)

### b) Upsampling a low dimensional noise:

Considering the above example, we could sample a 4×4 noise matrix and then upsample it to the input image size. However, the generator creates a noisy target image during the test time, as shown in Figure 1(b). Most of the time, the generator fails to make meaningful target domain images. It could be that the noise information is still high for the generator to treat it as a code, not a feature vector. Thus to not further limit the input noise vector dimensions, we defined the next scenario.

**c) Use a narrower bottleneck while feeding an upsampled low-dimensional noise:**
In this case, we add a module to the middle of the U-Net. This module downsamples the encoder's extracted features. Using such a module will lead the generator to treat the noise as a code similar to GANs and does not extract features from the noise, as shown in Figure 1(d). To reduce the network computation complexity and ensure that the network does not extract features from noise input, we can freeze the encoder part of the generator during the training procedure. Then the generator almost perfectly learns the target domain, as shown in Figure 1(e). However, if we only use the narrower bottleneck while using high dimensional noise, the skip connections in the U-Net architecture pass most of the noise information. Hence, the bottleneck becomes ineffective in helping the network learn the target distribution, as shown in Figure 1(c).

To conclude the above scenarios, we can change the U-Net architecture to play the role of a generator in a simple GAN model. We want to generate all possible outputs in the target distribution to feed them as fake images to the discriminator. A wide distribution of inputs helps the discriminator to backpropagate broad distribution gradients to the generator. Hence, we need to use a dynamic network to propose an architecture that can handle a high-dimensional image and low-dimensional noise input at the same time.

Based on the scenarios explained here, we propose a method that can condition the outputs on the input images. Simultaneously, we generate outputs based on the noise to help the generator learn the target domain accurately. This will result in the constructed outputs fitting better into the target domain distribution.

## 1.2 Related Works

In this section, we introduce the concepts that we have used in this research. Dynamic neural networks help us change the Pix2Pix architecture so that it can simultaneously process noise and real images to learn the target domain distribution thoroughly. We also studied the noise usage in the GANs and cGANs architecture to further understand the noise impact on the GANs and cGANs performance and abilities.

Dynamic neural networks change the network's architecture during the training or test phases. These changes are based on the input or information about the model's overfitting or backpropagated gradients. In most dynamic networks, changing the architecture during the training or testing procedure improves the model's performance based on the situation the model must handle [12].

GANs are powerful models for learning complex distributions to synthesize semantically meaningful samples [13]. The models in this field, such as [7-11], take a noise vector as input and attempt to learn the target distribution by changing the noise distribution with a decoder-like architecture. Noise, in general, is one of the critical components which helps GANs learn the target distribution. The distribution, dimensions, noise injection method, and the way that the network processes the noise affect the generalization, diversity, and quality of synthesized images [14].

Conditional Generative Adversarial Networks (cGANs) such as [2, 15-22] use the power of GANs to learn the correlations between the distributions of two or multiple domains of images and translate the images between these domains. One of the most famous models in this field is Pix2Pix [2]. This model is powerful in learning the correlations between source and target domain images. The Pix2Pix model attempts to use noise to increase the model's generalizability. However, the noise in their architecture has the same effect as the dropout layer in neural network architecture. Therefore, the noise component seems redundant in Pix2Pix and has a shallow impact on the output results; thus, it is empirically omitted from the cGAN's architecture. There are multiple other reasons that the noise is injected or processed into the cGANs or GANs architectures which could be classified as below:

**a) Noise robust architecture:**

In part of these works, the researchers used noise to make a robust architecture against noisy inputs [23-27]. For example, in [23], the model augments the generator in an unsupervised way, promoting the generator's outputs to span the target manifold even in the presence of intense noise.

**b) Learning distribution from noisy images:**

The other approach is to use the noise to neutralize its effect when we want to learn target distribution from noisy images [28].

**c) Many-to-many translation:**

One of the recent usages of the noise in cGAN architectures is to use it when we want a many-to-many translation [29-32]. In other words, in part of the works, such as translating sketched drawings to real-like images, it may be favorable to have multiple options in the network's output. For example, in fashion or interior design tasks, the user should have various options; hence, a stochastic architecture that can generate different outcomes for the same input image seems to be a need [31].

**d) Denoising or modeling the noise:**

Some other works in this field attempt to denoise input images [33-35]; thus, they need to model the noise [36, 37]. Modeling the noise of devices with GANs or cGANs architectures will become one of the hot topics due to the power of these models in learning distributions.

In this work, with the help of dynamic neural networks, we propose another application for the noise in the cGANs architecture, which helps this architecture learn a complete interpretation of target distribution like GANs but based on limited data. More precisely, we changed the Pix2Pix architecture in a way that could process noise and the real input image. This helps our model to learn the target distribution based on the noise while considering the input and target correlations such as normal Pix2Pix models.

This paper is organized as follows: Section 2 explains our dynamic network architecture and discusses how to process the noise and real images simultaneously. Then, in Section 3, we describe our implementations and experiments details and present obtained results. Finally, in Section 4, we discuss the cause of the obtained results.

## 2. Proposed Methods

Conditional GANs cannot generate all the possible fake images in the target domain. Therefore, backpropagated gradients from the discriminator (based on an incomplete training set) do not guarantee that the generator does not violate the target domain distributions. Suppose we want to be sure that the generator realizes all the critical features of the target domain images. Hence, we need to make all the possible output in the target domain, similar to GAN's architectures which learn target distribution by utilizing noise code as input. As shown in experiments that have been done in Section 1, feeding the noise to U-Net-like architectures like the generator of cGANs is not trivial. Therefore, we must inject the noise into the architecture so that the generator does not treat it like the input images, which will lead to mislearning in the generator training procedure. Figure 2 presents a block diagram of our proposed method, in which we show the learning procedure from an input image and a noise input. A critical point here is that we train the network based on image and noise in two different cycles by dynamically changing the network's bottleneck during the training. Once the input is a real image, the network passes it through the black path, and in the noise injection cycle, the network passes the noise through the yellow path. We will explain these two learning cycle procedures in the upcoming subsections.

### 2.1 Learning from Real Input Images

We feed the input image to the generator, similar to regular cGAN architectures, and the generator conditions the output on the features extracted by the encoder. Then, similar to regular cGAN's gradients update cycle, we first freeze the generator to train the discriminator. The generator output is concatenated with the input image and passed through the discriminator as the fake image pair. The real target label and input image are also concatenated and fed to the discriminator as the real image pair.

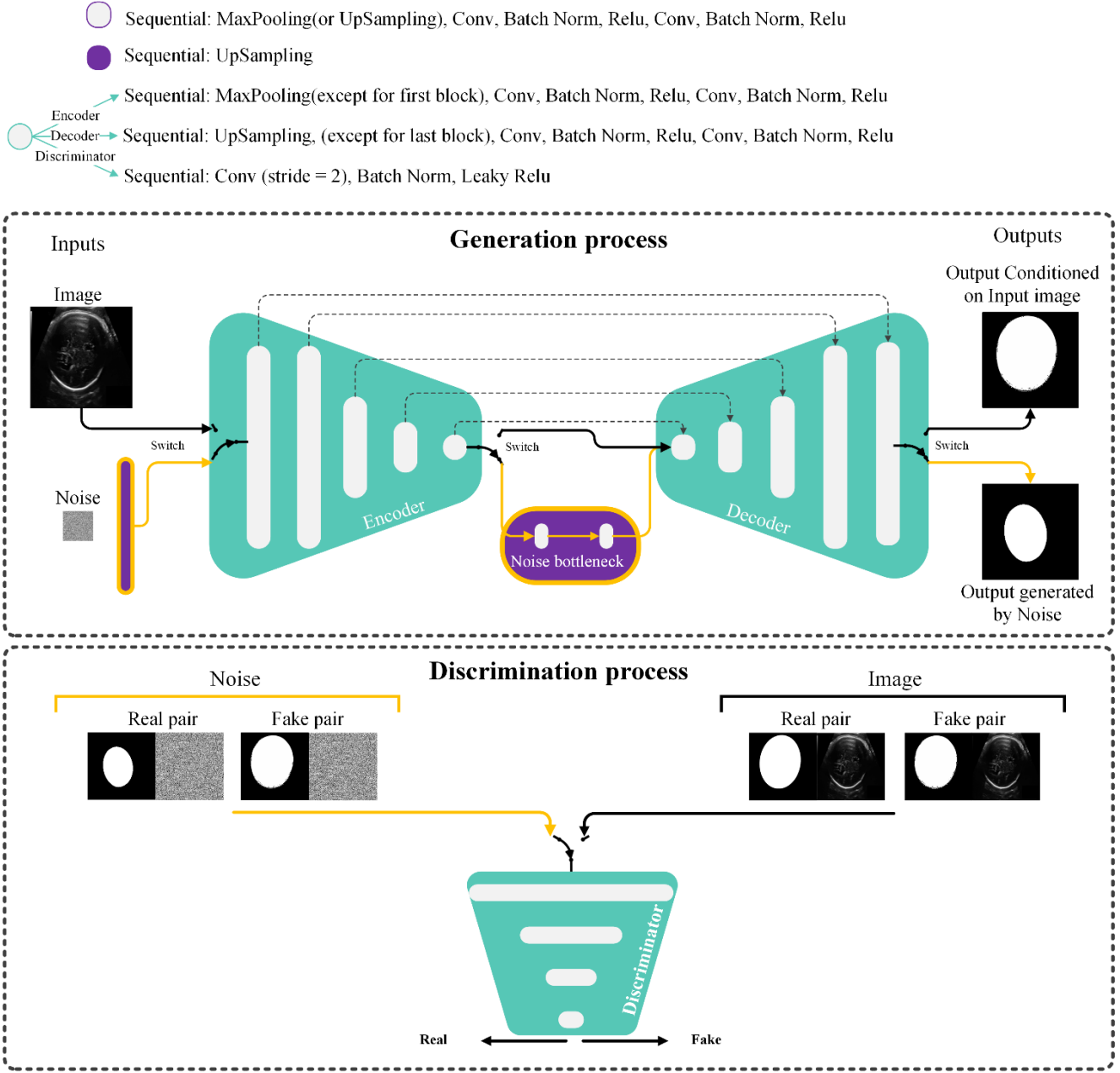

Figure 2: Block diagram of our proposed method. As we can see, we pass the Image and the Noise into the network from the black and yellow paths, respectively. The noise bottleneck works when the input is noise.

Let us assume $G$, $D$, $Image$, represent the generator, discriminator, and input image. Also, $Output_{\text{Image}}$, represent the generator's output when the input is an image. Furthermore, $P_{data}$ and $P_G$ represent the distributions of ground truth images and the generated images, respectively. The discriminator loss is calculated based on Equation 1. Then, the gradient of the loss is used to update the discriminator's learnable parameters to distinguish between the generated and ground truth target images.

$$\mathcal{L}_{D_{Image}}(D, Ground\ truth, Output_{Image}) = -\frac{1}{2}\mathbb{E}_{Groud\ truth \sim P_{data}}(\log(D(Groud\ truth) - 1)) \\ -\frac{1}{2}\mathbb{E}_{Output_{Image} \sim P_G}\left(\log\left(D(Output_{Image})\right)\right) \quad (1)$$

In the above equation, the $D(Groud\ truth)$ and $D(Output_{Image})$ are the output of the discriminator for the ground truth and the output of the generator for the input images, respectively. On the other hand, according to the discriminator output, we calculate the loss of the generator based on Equation 2. Then we freeze the discriminator to update the generator parameters to generate more realistic images. The discriminator should not distinguish these images from real target images.

$$\mathcal{L}_{G_{D_{Imag}}}(G, D, Image) = \frac{1}{2}\mathbb{E}_{Image \sim P_{data}}(\log(D(G(Image) - 1))) \quad (2)$$

We also calculate the L1 distance loss between the generated output and ground truth target image based on Equation 3, similar to the Pix2Pix model [2], to further guide the generation procedure to construct exact details.

$$\mathcal{L}_{L1_{Image}}(G, Ground\ truth, Image) = \mathbb{E}_{Ground\ truth,\ Output_{Image}} ||Ground\ truth - G(Image)|| \quad (3)$$

Notice that the $G(Image)$ and $Output_{Image}$ both show generator's output. However, $G(Image)$ presents the output, which the computed losses based on that only update generator's parameters. Furthermore, $Output_{Image}$ is detached from the generator and is used to calculate the discriminator's loss and update the discriminator's parameters.

## 2.2 Learning from Noise Input

To pass the noise input through the generator, we sample a 4×4 matrix from a uniform distribution and then feed it to an upsampling layer to generate an array that has the same size as an input image. Hence, the noise elements become highly correlated after upsampling and do not present too much information to the network. We pass this generated noise image through the generator using the yellow path, as shown in Figure 2. We use a noise bottleneck module to narrower the network and not let the network pass too much noise information to the decoder part of the generator. This bottleneck highly limits the number of channels and also uses a max-pooling layer to filter further the features that the noise input could pass to the decoder (the shape of noise bottleneck module output is 1×4×4). Therefore, we feed a low dimensional noise to the decoder, producing an output that the discriminator tries to distinguish from real images. The model does not use L1 distance loss for noise-generated outputs. It is worth mentioning that the loss calculated in the discriminator's output for noise input does not update the encoder's parameters because we do not want the network to extract features from the noise input. The discriminator's and generator's loss for noise input is computed based on Equations 4 and 5. The encoder processes the noise and then feeds it with multiple connections to the decoder part. This procedure is like how StyleGAN [9] treats the input noise vector.

$$\mathcal{L}_{Dis_{Noise}}(D, Output_{Noise}, Groun\ truth) = -\frac{1}{2}\mathbb{E}_{Groud\ truth \sim P_{data}}(\log(D(Groud\ truth) - 1)) \\ -\frac{1}{2}\mathbb{E}_{Output_{Noise} \sim P_G}(\log(D(Output_{Noise}) - 0)) \quad (4)$$

$$\mathcal{L}_{G_{D_{Noise}}}(G, D, Noise) = \frac{1}{2}\mathbb{E}_{Noise \sim Uniform[-1,\ 1]}(\log(D(G(Noise) - 1))) \quad (5)$$

Where the $G$, $D$, $Noise$ and $Output_{Noise}$ stand for the generator, discriminator, input noise, and the output generated by the discriminator when the input is noise.

The total learning procedure is modeled mathematically as a minimax game between the generator and discriminator, presented in Equation 6.

$$G^* = \arg\min_G \max_D \alpha \times \mathcal{L}_{L1_{\text{Image}}}(G, \text{Ground truth}, \text{Image})$$
$$+ \beta \times \mathcal{L}_{G_{D_{\text{Image}}}}(G, D, \text{Image}) \quad (6)$$
$$+ \beta \times \mathcal{L}_{G_{D_{\text{Noise}}}}(G, D, \text{Noise})$$

Where $\mathcal{L}_{G_{D_{\text{Noise}}}}$ presents the loss of a generated image using an input noise detected by the discriminator. Furthermore, $\mathcal{L}_{L1_{\text{Image}}}$ and $\mathcal{L}_{G_{D_{\text{Image}}}}$ are the loss of a generated image using an input image that has been calculated using the L1 norm and the loss of generated images using an input image detected by the discriminator, respectively. The α and β also present the importance of generator losses based on the L1 norm and the discriminator's output, respectively. We considered $\alpha = 10$ and $\beta = 1$ for all of the experiments.

Notice that the $G(Noise)$ and $Output_{\text{Noise}}$ both show the generated output by the generator. However, the first one presents the output, which the losses computed based on that only update generator parameters. The second one is detached from the generator to compute the loss for the discriminator and only updates the parameters.

## 3. Experimental Results

In this section, we separate experimental results and setups into two parts. First, we compare our model abilities with in-domain segmentation tasks. Then we compare our model abilities for out-domain generalization tasks. Next, we explain the datasets in each part of the experiments in a subsection. Then we describe the model's hardware and hyperparameters in the Implementation Details subsection. Next, in the Result subsection, we will discuss the quantitative and qualitative results and the training curves of the models.

### 3.1 In-Domain Segmentation Ability

#### 3.1.1 Datasets

We use two different datasets to make a comprehensive comparison. We chose the segmentation task to compare our proposed model with the Pix2Pix model because the segmentation task can be numerically evaluated and shows the model's power better. The HC18 [38] dataset contains 999 ultrasound images of the fetal head. The Montgomery chest x-ray [39] dataset has 114 images. We use these datasets to compare our model in domain segmentation power against Pix2Pix and U-Net. We first separate the training, validation, and test sets into the proportion of 70, 10, and 20 percent. Then we resize the images to 288×288 and crop them to 256×256.

#### 3.1.2 Implementation Details

We implement our models using the PyTorch platform. We train our implementations using an Nvidia Geforce GTX 1080 ti. We set the initial learning rate to 0.0002 for all models to have a fair comparison among them. The HC18 dataset needs 200 epochs to converge. The model is trained with a constant learning rate in the first 100 epochs, and in the second 100 epochs, the learning rate linearly decreases to zero. On the other hand, the Montgomery dataset needs 50 epochs to converge, and the training in the first 30 epochs is performed with a constant learning rate. Then in the remaining epochs, we linearly decrease the learning rate toward zero.

#### 3.1.3 Results

Figure 3 presents the training curves for our model on both datasets. As we can see, the generator's L1 loss presented in the plots by orange color converged after the number of epochs increased. The discriminator losses, shown in green and red colors, converge after the determined number of epochs.

However, the discriminator's convergence causes significant generator losses. These losses are computed based on the discriminator output, as shown by the blue curve. The generator tries to decrease this loss, but because the L1 loss has a higher coefficient ($\alpha$) than the discriminator loss coefficient ($\beta$), the generator prioritizes the L1 loss.

As a result, the blue curve does not decay to zero. However, it does not mean that the generator does not learn the desired target distribution. Instead, the blue curve indicates that the discriminator's distinguishing power grows.

As shown in Figure 4, the generator learns the target distribution based on the uniform noise fed to the generator as the input. This learning guarantees that the decoder produces outputs that fit the target distribution with or without attention to the input image features. The generator does not sacrifice fitting to the target domain distribution to correlate the output more with the input information. In other words, the encoder extracts the features from the input image, which helps the decoder to construct an output precisely in the target distribution.

We can also consider that our generator can learn a complete interpretation of the target distribution, similar to GANs. Each latent space code we feed to the decoder presents a target domain image. However, some latent code features extracted from real input (not noise) represent a specific target domain image that correlates to the real input image. Therefore, in the test time, because the latent code is generated from real input, the generator constructs an output in the target domain that correlates with the input image. The obligation we made for the generator by using noise will result in more realistic

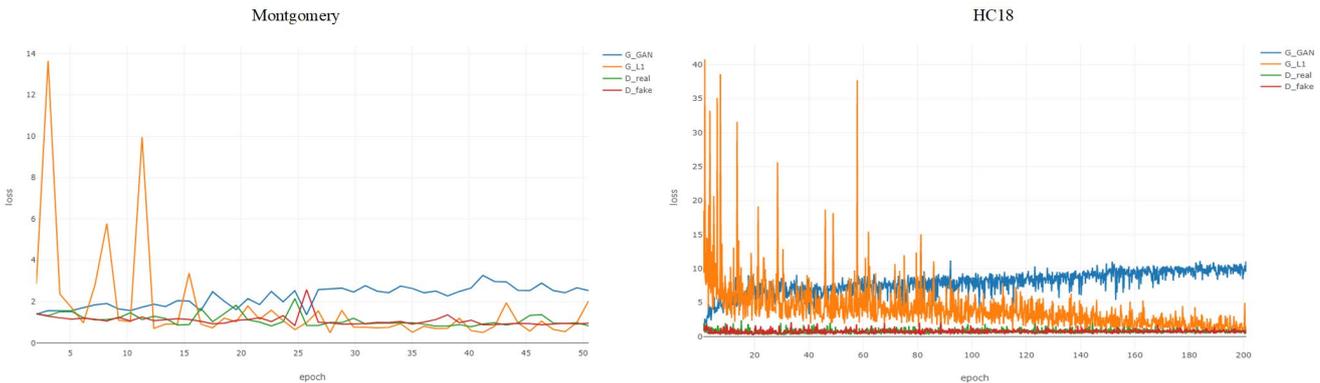

Figure 3: Training curves of our proposed model using Montgomery and HC18 datasets.

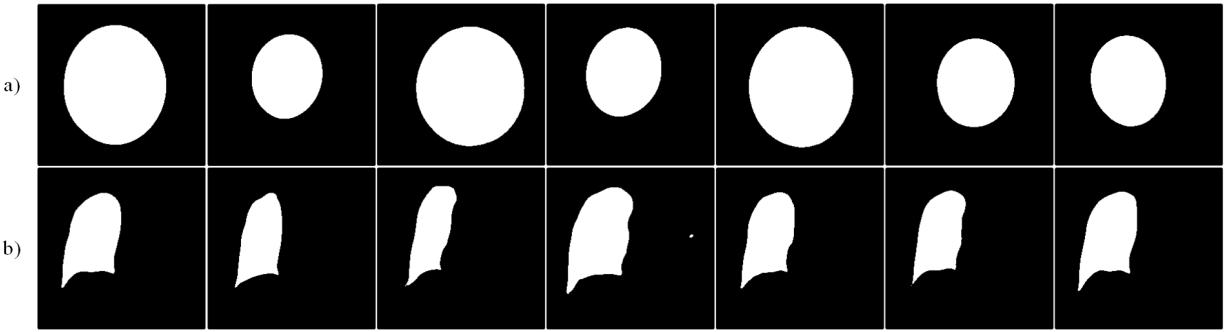

Figure 4: Generated outputs using noise input. The first and the second row show the generated target domain images based on the noise using the a) HC18 and b) Montgomery datasets, respectively.

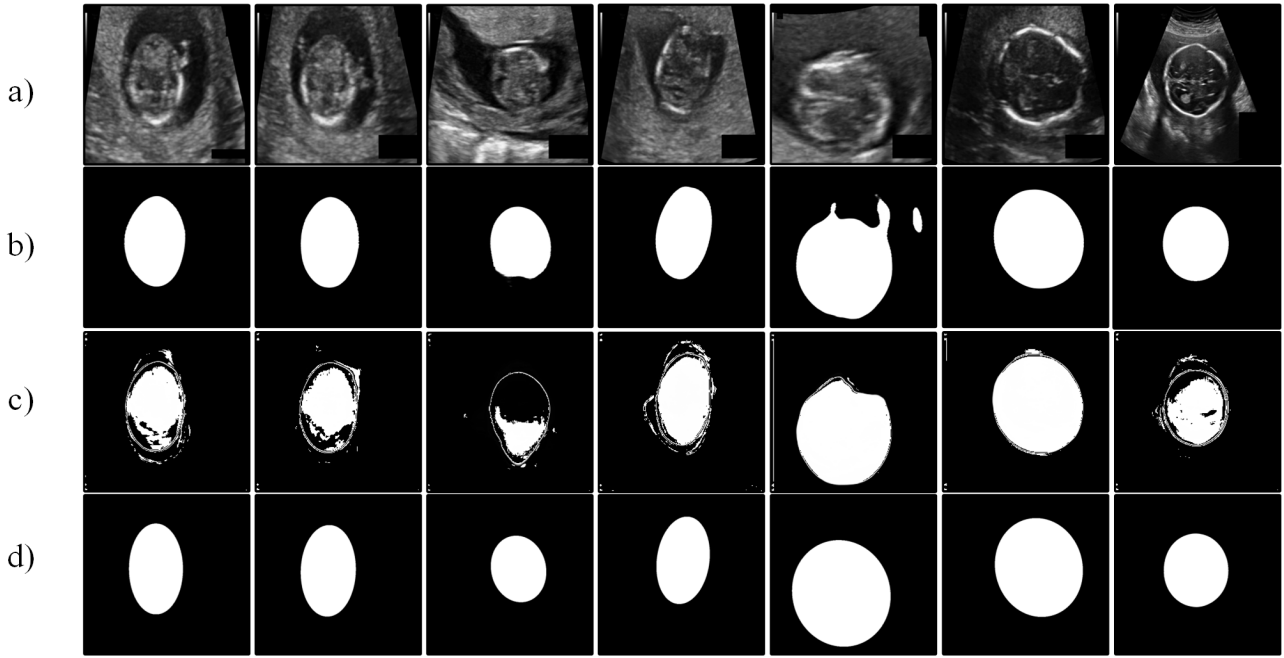

Figure 5: The results of our proposed model versus Pix2Pix on both HC18 datasets: a) Input image b) generated outputs by using our proposed model c) generated outputs using Pix2Pix model d) ground truth

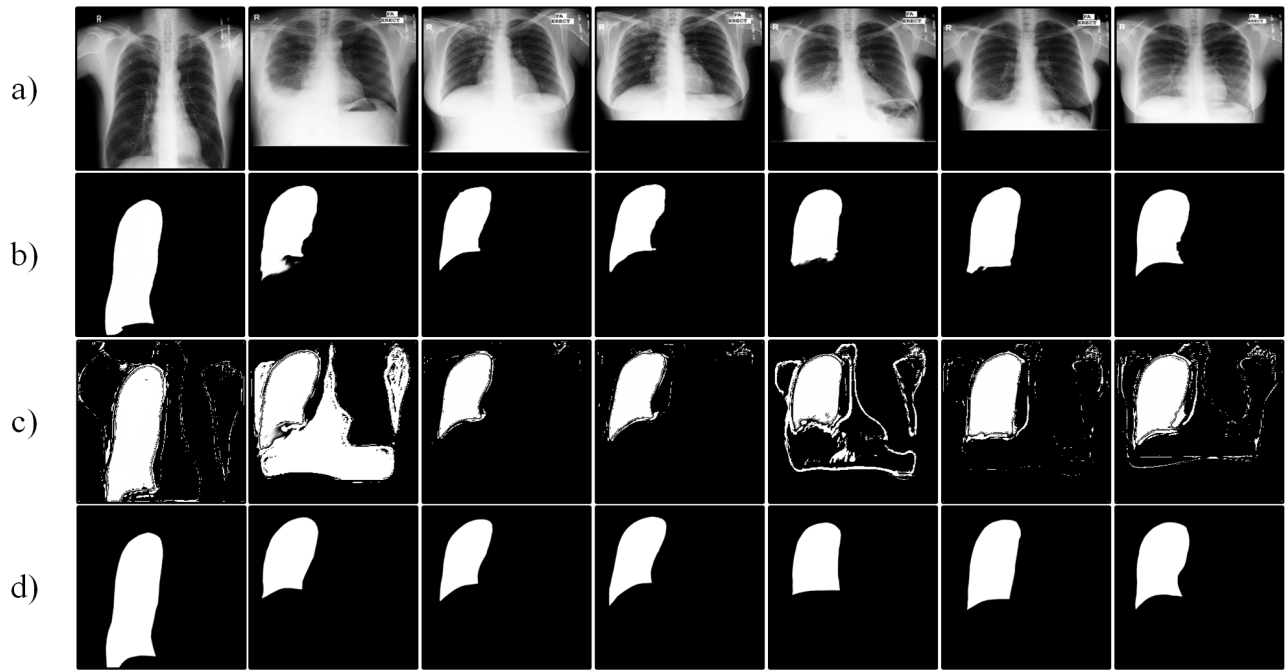

Figure 6: The results of our proposed model versus Pix2Pix on Montgomery datasets: a) Input image b) generated outputs by using our proposed model c) generated outputs using Pix2Pix model d) ground truth

images in the target domain distribution, as shown in Figure 5 and Figure 6. Table 1 compares our method with the Pix2Pix method based on the Dice score of generated masks during the test. As we expected, our network segments more realistic and accurate masks for each image during the test, outperforming the Pix2Pix method. We did not use any augmentation method because we wanted a fair comparison, and the results were only achieved based on the original training set.

Table 1: Comparison of our proposed method versus the Pix2Pix method on both HC18 and Mongomery x-ray datasets using Dice scores.

| Dataset | Method | Dice score (%) |
|---|---|---|
| HC18 | Pix2Pix | 91.86 |
| | Our proposed method | **97.28** |
| Montgomery | Pix2Pix | 82.95 |
| | Our proposed method | **97.29** |

### 3.2 Out-Domain Segmentation Ability

Our method in the yellow path, as shown in Figure 2, treats every input as noise and does not extract features from the image. Furthermore, as we wanted, the generator's decoder, regardless of the input images, modeled the target domain distribution (ultrasound fetal head shape) almost perfectly. This will help the model generalize better on out-domain images. Finally, we investigate our model generalizability by using the below settings, and we will discuss the situations in which our model can generalize better.

#### 3.2.1 Datasets

We trained our model on JSRT [40] chest x-ray and ISIC [43] skin melanoma datasets. The first task is to segment lungs from x-ray images, and the second is to segment skin lesions. When we trained our model with JSRT, we tested our model on NLM [41], NIH [42], and SZ [41] chest x-ray sets as out-domain datasets. We also tested the model, which was trained with the ISIC dataset with PH2 [44], IS [45], and Quest [45] skin lesions datasets. To compare our models in a fair scenario, we set the initial learning rate to 0.0002 for all models.

#### 3.2.2 Implementation Details

The implementation hardware and platform are similar to section 1.1.2. Both JSRT and ISIC datasets need 200 epochs to converge, which the model trained with constant learning in the first 100 epochs, and in the second 100 epochs, the learning rate linearly decreased to zero.

#### 3.2.3 Results

As we can see in Table 2, our model achieves comparable results with semantic GAN [3] on the JSRT dataset. However, in Table 3, although our model shows better generalization in most cases against other models, its weaknesses against semantic GAN become clearer. In semantic GAN, to learn the joint target and input domain distribution, thousands of images are used to train the StyleGAN v2 [9].

Table 2: Chest X-ray Lung Segmentation. Numbers are DICE scores. JSRT [40] is the in-domain dataset on which we train and evaluate. We also assess additional out-of-domain datasets (NLM [41], NIH [42], SZ [41]). SemanticGAN [3] used 108k other unlabeled images to learn the distribution (Red cells: best, Blue cells: second best)

| Method \| test set | Trained with 9 labeled data samples | | | | Trained with 35 labeled data samples | | | | Trained with 175 labeled data samples | | | |
|---|---|---|---|---|---|---|---|---|---|---|---|---|
| | JSRT | NLM | NIH | SZ | JSRT | NLM | NIH | SZ | JSRT | NLM | NIH | SZ |
| U-Net [4] | 0.9318 | 0.8605 | 0.6801 | 0.9051 | 0.9308 | 0.8591 | 0.7363 | 0.8486 | 0.9464 | 0.9143 | 0.7553 | 0.9005 |
| DeepLab [46] | 0.9006 | 0.6324 | 0.7361 | 0.8124 | 0.9556 | 0.8323 | 0.8099 | 0.9138 | 0.9600 | 0.8175 | 0.8093 | 0.9312 |
| MT [47] | 0.9239 | 0.8287 | 0.7280 | 0.8847 | 0.9436 | 0.8239 | 0.7305 | 0.8306 | 0.9604 | 0.8626 | 0.7893 | 0.8846 |
| AdvSSL [48] | 0.9328 | 0.8500 | 0.7720 | 0.8901 | 0.9552 | 0.8191 | 0.5298 | 0.8968 | 0.9684 | 0.8344 | 0.7627 | 0.8846 |
| GCT [49] | 0.9235 | 0.6804 | 0.6731 | 0.8665 | 0.9502 | 0.8327 | 0.7527 | 0.9184 | 0.9644 | 0.8683 | 0.7981 | 0.9393 |
| SemanticGAN [3] | 0.9591 | 0.9464 | 0.9133 | 0.9362 | 0.9668 | 0.9606 | 0.9322 | 0.9485 | 0.9669 | 0.9509 | 0.9294 | 0.9469 |
| Ours | 0.9609 | 0.8976 | 0.8526 | 0.8620 | 0.9717 | 0.9236 | 0.9230 | 0.9221 | 0.9778 | 0.9387 | 0.9145 | 0.9349 |

Table 3: Skin Lesion Segmentation. Numbers are JC index. ISIC [43] is the in-domain dataset on which we train and evaluate. Additionally, we perform segmentation on three out-of-domain datasets (PH2 [44], IS [45], Quest [45]). SemanticGAN [3] used 33k additional unlabeled images to learn the distribution. (Red cells: best, Blue cells: second best)

|  | Trained with 40 labeled data samples | | | | Trained with 200 labeled data samples | | | | Trained with 2000 labeled data samples | | | |
|---|---|---|---|---|---|---|---|---|---|---|---|---|
| Method \| test set | ISIC | PH2 | IS | Quest | ISIC | PH2 | IS | Quest | ISIC | PH2 | IS | Quest |
| U-Net [41] | 0.4935 | 0.4973 | 0.3321 | 0.0921 | 0.6041 | 0.7082 | 0.4922 | 0.1916 | 0.6469 | 0.6761 | 0.5497 | 0.3278 |
| DeepLab [46] | 0.5846 | 0.6794 | 0.5136 | 0.1816 | 0.6962 | 0.7617 | 0.6565 | 0.4664 | 0.7845 | 0.8080 | 0.7222 | 0.6457 |
| MT [47] | 0.5200 | 0.5813 | 0.4283 | 0.1307 | 0.7052 | 0.7922 | 0.6330 | 0.4149 | 0.7741 | 0.8156 | 0.6611 | 0.5816 |
| AdvSSL [48] | 0.5016 | 0.5275 | 0.5575 | 0.1741 | 0.6657 | 0.7492 | 0.6087 | 0.3281 | 0.7388 | 0.7351 | 0.6821 | 0.6178 |
| GCT [49] | 0.4759 | 0.4781 | 0.5436 | 0.1611 | 0.6814 | 0.7536 | 0.6586 | 0.3109 | 0.7887 | 0.8248 | 0.7104 | 0.5681 |
| SemanticGAN [3] | 0.7144 | 0.7950 | 0.7350 | 0.5658 | 0.7555 | 0.8154 | 0.7388 | 0.6958 | 0.7890 | 0.8329 | 0.7436 | 0.6819 |
| Ours | 0.6517 | 0.6693 | 0.5821 | 0.3047 | 0.7060 | 0.7521 | 0.6826 | 0.5632 | 0.7669 | 0.8021 | 0.6939 | 0.6321 |

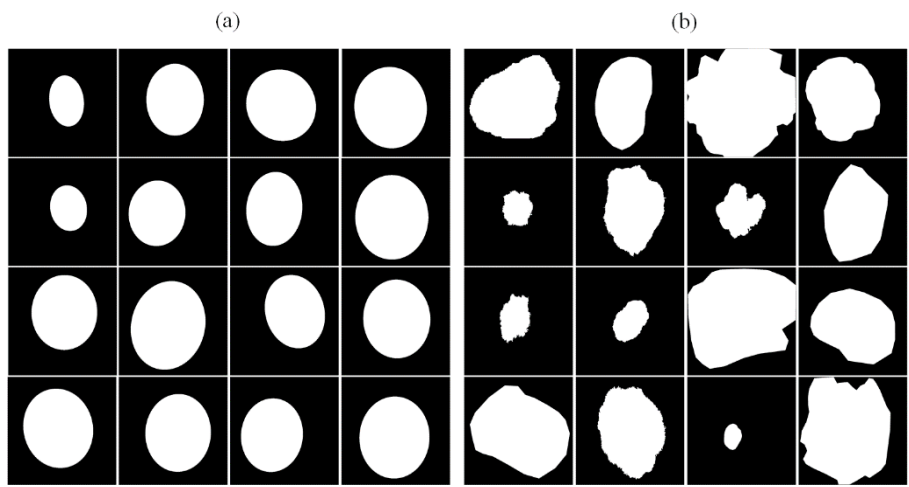

Figure 7: Some of the ground truth maps from a) HC18 and b) ISIC datasets.

Table 4: Effects of different latent sizes on VAE performance to learn the HC18 and ISIC ground truth distribution.

| | Dice | |
|---|---|---|
| Latent size | HC18 | ISIC |
| 2 | 84.06 | 70.96 |
| 3 | 90.63 | 81.32 |
| 4 | 96.10 | 88.02 |
| 8 | 96.16 | 92.79 |
| 16 | 96.15 | 87.30 |
| 32 | 96.18 | 70.50 |

Then they used an encoder to encode the images to be segmented and fed to the generator. Therefore, their network can learn a more comprehensive input and output distribution description. However, in most medical image segmentation tasks, we cannot have thousands of images, which could be the bottleneck of the method of reference [3]. Besides, training such a model is too time-consuming. Therefore, the question is, in which situation could our model generalize almost similar to semantic GAN?

To answer this question, we trained a simple Variational Auto Encoder (VAE) model multiple times with different latent sizes on HC18 and ISIC segmentation ground truths, as shown in Table 4. We can see that HC18 ground truth space could be mapped almost perfectly to four-dimensional space. This is because all the labels have an ellipse shape, as shown in Figure 7(a), and the distribution is compactable. Thus learning the distribution with a low number of samples is possible. But in the ISIC, the ground truth could have any shape, as shown in Figure 7(b). Thus, modeling it in low dimensions is almost impossible. Increasing the latent size also could not help because the number of samples is limited.

To conclude the above explanation, if we have a segmentation task in which the labels have a shape, for example, lunge images in the Montgomery dataset or fetal head shape in the HC18 set, our model could generalize perfectly. For instance, we want to extract elliptical shapes when segmenting ultrasound images to measure the fetal head circumference. However, if the ground truth masks have a vast distribution, we must have large datasets and use other architectures such as semantic GAN. Using such architectures is time-consuming, needs powerful hardware, and requires large datasets.

## 4. Conclusion

In this work, we proposed an architecture that combined characteristics of the GAN and conditional GAN models. Using the GANs characteristics, which learn the target distributions from noise inputs in the cGANs architecture, leads cGAN to learn the target distribution better than the native cGAN architecture. Our architecture has lower complexity than state-of-the-art methods. On HC18 and Montgomery in-domain datasets, we did not use image augmentation methods, and our network shows comparable results. Furthermore, input noise in our network helped the model generalize better on unseen data. Our model showed strong out-domain generalization for data labels that have a pattern. Therefore, this architecture can produce outstanding results on small datasets. Our research could be continued by applying this technique to a cycleGAN [50] framework, which needs a complete interpretation of the target and source domain more than cGANs.